# Towards Humanoid Robot Autonomy: A Dynamic Architecture Integrating Continuous thought Machines (CTM) and Model Context Protocol (MCP)

Libo Wang
UCSI University
Nicolaus Copernicus University
free.equality.anyone@gmail.com / 1002265630@ucsiuniversity.edu.my

*Abstract*— **To address the gaps between the static pre-set "thinking-planning-action" of humanoid robots in unfamiliar scenarios and the highly programmed "call tool-return result" due to the lack of autonomous coding capabilities, this work designs a dynamic architecture connecting continuous thought machines (CTM) and model context protocol (MCP). It proposes a theoretical parallel solution through tick-slab and uses rank compression to achieve parameter suppression to provide a solution for achieving autonomous actions due to autonomous coding. The researcher used a simulation-based experiment using OpenAI's o4-mini-high as a tool to build the experimental environment, and introduced the extended SayCan dataset to conduct nine epochs of experiments. The experimental results show that the CTM-MCP architecture is feasible and effective through the data results of seven metrics: task success rate (TSR), execution success rate (ESR), average episode length (AEL), ROSCOE, REVEAL, proficiency self-assessment (PSA), task effectiveness (TE). In practice, it provides a reference experience for exploring the autonomous dynamic coding of humanoid robots based on continuous thinking to achieve human-like autonomous actions.**

## I. Introduction

As large language models (LLMs) continue to expand towards multimodal interaction and autonomous operation, the model context protocol (MCP) has gradually become an important intermediary framework to promote the practical application of AGI as a bridge between serial semantic generation and external tool calls (Hou et al., 2025). In view of the need for the integration of multimodal AI and control machine learning, especially from the exploration of Constitutional AI and tool-use orchestration by institutions such as Anthropic and OpenAI, it enables the model to have a clear and transferable encapsulation protocol when processing context extension and instruction deconstruction (Anthropic, 2022; OpenAI, 2025).

MCP proposed by Anthropic (2024) is essentially an open standard that enables developers to build secure bidirectional connections between data sources and AI-driven tools (Fig. 1). It aims to promote the consistency of communication protocols between LLMs external data sources and tools. It is committed to addressing the limitations of models due to data silos, so that AI agents can access and operate local and remote data more securely (Yang et al., 2025). The bidirectional connection provides an interface to connect everything, which enables developers to expose data through the MCP server and more conveniently operate multiple tools such as applications connected to the server (Narajala & Habler, 2025).

Based on the open standard of JSON-RPC 2.0 architecture, the key technical core of MCP is to abstract external tools into a schema-driven remote procedure call interface and realize the automation of instruction scheduling through the tools protocol (Ehtesham et al., 2025). It uses the internal implementation logic including a two-way pairing mechanism of capability declaration and structured arguments to enable AI agents to dynamically generate function calls based on context without pre-hard coding (Krishnan, 2025).

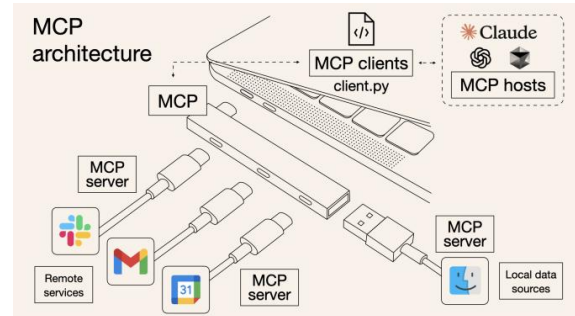

Fig. 1    MCP Architecture (Adapted from Sakal, 2025)

From an application perspective, MCP promotes generative AI to become autonomous agents with the ability to execute operations and self-manage tasks that can self-decompose subtasks, select appropriate tools, call execution, and update strategies based on feedback (Habler et al., 2025). As technology matures and is widely used, it increasingly inspires users to realize the embodiment of humanoid robots that realize autonomous perception, decision-making, planning and action close to that of humans. In light of vision and purpose, humanoid robots present different development axes. Electronic skeletal systems represented by Tesla emphasize high-frequency response and integrated control to complete specific repetitive tasks in factory scenarios; while musculoskeletal humanoids represented by Clone Robotics pursue the faithful reproduction of human movement semantics and tactile feedback (Fig. 2).

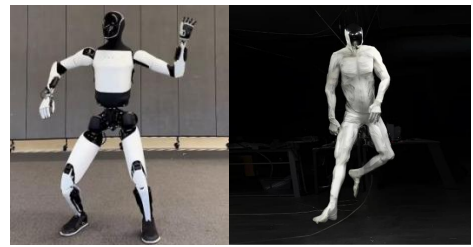

Fig. 2    Tesla's Optimus and Clone Robotics' Protoclone

However, video footage released by many companies and research institutions shows a common problem: current humanoid robots are unable to demonstrate autonomous "thinking-planning-acting" capabilities close to those of humans, and thus can only perform single functions and special purposes (Spong, 2022; Li et al., 2024). The fundamental gap that makes it difficult to address is that the control systems of most humanoid robots use pre-arranged rather than dynamically generated solutions to tasks (Bärmann et al., 2024). It is not based on real-time scene understanding and strategy construction, but is designed by AI engineers through static coding or remote operation of preset parameters, and embedded in the control system through structures such as behavior tree, scripting modules or state transition graph (Ghzouli et al., 2023). Although repetitive actions have been gradually reduced, whether it is moving objects, dancing, assembly work or even boxing competitions, the humanoid robot has not yet demonstrated dynamic autonomous encoding through perception in unknown environments in practice (Bottega et al., 2025). It has led to the formation of a closed loop architecture of "semantic recognition-strategy fixation-execution output" in humanoid robots, which directly limits its cross-scene adaptability and spontaneity of action. It is the deep reason why this work introduces MCP to enhance the ability of humanoid robots to respond to new tasks of immediate perception in complex environments.

The second problem is that MCP may form a highly programmed "call tool-return result" interaction mode in the process of applying it to humanoid robot control system to realize autonomous coding, autonomous planning and autonomous action. It is a control system of embodied intelligent robot that degenerates into a simple mechanical mode due to lack of life cycle management in execution, which is manifested as a rigid combination of functions like a stitching monster. Although the protocol has characteristics of interface universality and simplified scheduling, it fails to provide integrated support for the compilation process of dynamically generated code, static analysis mechanism, sandbox execution protection and error feedback (Hou et al., 2025; Krishnan, 2025). It makes it difficult for MCP to meet the stability requirements of error detection, action verification and version rollback in continuous operation and real-time control in multiple scenarios in the process of realizing humanoid robot autonomy. The fundamental gap that causes the above problem comes from the lack of a thinking module similar to the human brain's coherent strategic reasoning and semantic constraint capabilities before the MCP call process. It is committed to task-context resolution, tool selection reasoning, parameter synthesis rules and error causality forecasting as a structured thinking layer, so as to build autonomy from task semantics to tool scheduling.

In view of the above gaps, the researcher introduced and improved the continuous thought machines (CTM) architecture proposed by Darlow et al (2025) and connected it with MCP to realize the current humanoid robot's autonomous and dynamic coding, thinking, planning and action scheduling from a principle perspective.

## II. Related work

As evidenced by the work of Darlow et al. (2025), the design of CTM deviates from the reasoning mode of traditional transformer models, which is mainly based on forward propagation and fixed attention mechanisms, and turns to simulating the internal temporal dynamics of neurons. This architecture uses neural synchronization and sequential memory evolution as the core logic to construct a potential structure with continuous thinking and adaptive computing capabilities. The key is that each neuron maintains an independent set of thought state records that include pre-activations, post-activations, synchrony weights and time decay parameters. It makes artificial neurons no longer static mapping units, but can form multiple rounds of evolution steps in the internal time dimension to perform repeated reasoning and strategy adjustments (Darlow et al., 2025).

In terms of design principle, the internal thinking process of CTM is built on recurrent inner loops. It allows the model to self-update the thinking state in a single input time step and adjust the temporal dependencies between different neurons through a learned decay function. The synchronization matrix, as a differentiable reasoning structure analogous to the biological neural phase synchrony, enables the model to learn context-coupled structures at different levels. In addition, CTM not only provides the ability of delayed memory, but also implements a dynamic computation controller embedded in the neural representation space that allows neurons to perform iterative structural search in internal time. Compared to the transformer model that relies solely on positional encoding and rigid layer configuration, CTM provides a neural execution model with temporal flexibility and computational density self-adjustment mechanism, which is particularly suitable for task scenarios that require multi-stage, hierarchical reasoning (Darlow et al., 2025).

Regarding the application and development of MCP technology, Hou et al. (2025) focused on the structural integration problem in the multimodal model architecture and proposed to semantically connect the language model with the external program logic in the form of a standardized intermediate layer. It is based on the MCP principle and focuses on establishing a three-level interaction system: prompt routing, tool binding layer, and MCP host scheduler. The core is to restructure the internal context output of the language model into a function node that meets the requirements of JSON-RPC calls, and to automatically match semantic tags and construct parameters for external APIs in the reasoning design capability declaration grammar. In terms of technical principles, this design breaks the limitation that traditional natural language input can only call static functions, and instead uses dynamic context to correspond to functional metadescriptors to form a dynamically combinable task graph. Hou et al. demonstrated that MCP, as a mediating protocol between models and external operation interfaces, has the potential to transform language models into multi-functional decision agents (Hou et al., 2025).

In addition, Krishnan (2025) conducted in-depth empirical research on the application of MCP in multi-agent systems to effectively address the problems of context retention and orchestration disconnection. The design of the MAS architecture combined with MCP was evaluated based on knowledge management and distributed decision-making tasks, and finally proved that MCP can share memory states across agents to improve continuous reasoning and context alignment capabilities (Krishnan, 2025). Narajala and Habler (2025) explored the enterprise-level security deployment of MCP and proposed a multi-layer defense strategy. It revealed that MCP may suffer from tool poisoning, command injection, and version rolling vulnerabilities in the tool scheduling process, and established a risk model with Zero Trust as the core (Narajala & Habler, 2025).

## III. ARCHITECTURE

The architecture proposed aims to prove the feasibility of connecting continuous thought machines (CTM) and model context protocol (MCP), so that humanoids simulate real-time human thinking in complex situations to realize autonomous coding execution (Fig. 3). It constructs a slab builder to concurrently generate k parallel reasoning branches in the synchronous control flow, and synchronizes the results that first pass the threshold through the consensus aggregator to achieve substantial parallelization under low-latency conditions. To control parameter inflation and computational overhead, it designed and introduced lightweight dynamic modules such as low-rank μ-MLP and syncrank1updater. It is committed to each reasoning branch share a low-rank parameter update mechanism to synchronously maintain thinking consistency and resource efficiency.

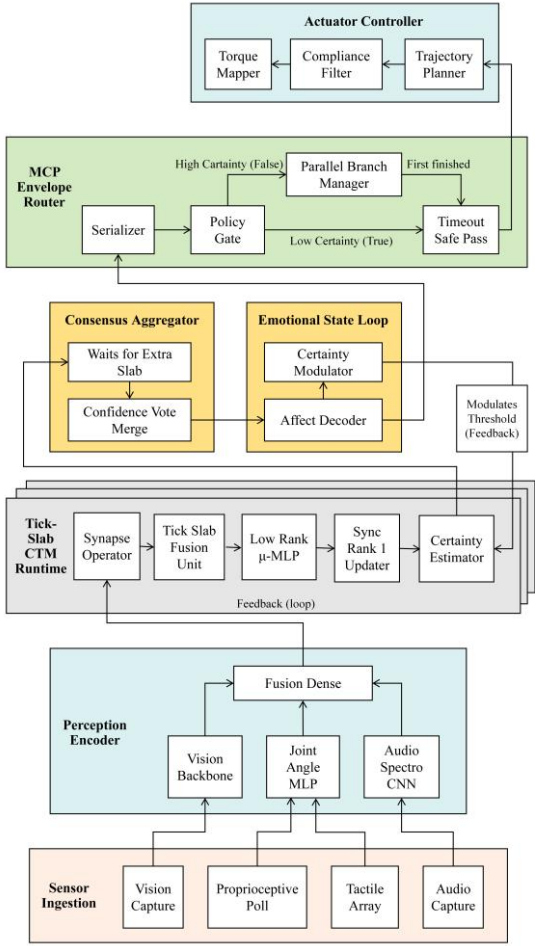

Fig.3 CTM-MCP Architecture

In the process of integrating the improved CTM architecture with the structural connection of MCP, after receiving instructions or sensing data, the humanoid robot no longer directly calls the tool but continuously tracks the task context and environmental variables through CTM to generate a acting chain through semantic reasoning and strategic planning. This architecture theoretically realizes the continuous closed-loop control mechanism of the robot's autonomous thinking, autonomous coding, and autonomous action. Its specific modules and algorithms are shown below.

### A. Sensor Ingestion

In the architecture, the sensor ingestion module, as the input layer of the control system, is responsible for the modal separation and preliminary encoding of the sensory data. It has the core function of maintaining the original semantic structure and corresponding to the downstream dedicated encoder. Each component of this module is directed to the corresponding dedicated next module according to its signal characteristics to form a directional data flow path. Among the components, the vision capture transmits the RGB-D data to the vision backbone through the image sequence processing flow. The design follows the principles of residual stacking and spatial tensor preservation to extract environmental structures, object boundaries, and action dependencies through multi-layer convolutions, while keeping the time dimension unfolded for subsequent timing module calls.

The signals of proprioceptive poll and tactile array belong to the body's internal sensing. The perception is semantically mapped to joint displacement, pressure distribution and end force estimation. Therefore, they are jointly transmitted to the joint angle MLP, and the space-action correspondence is encoded using nonlinear mapping and multi-layer hidden layers to form a semantic vector with posture structure connotation. In contrast, audio capture first performs short-time Fourier transform to map the original audio signal to spectral space, and then inputs it into Spectro CNN for convolution processing to retain the language features, rhythm changes and prompt signals implicit in the sound modality. The algorithm used in this module is as follows:

$$y = \tanh(W\, x)$$

where x represents the modality-specific input vector (pixels, mel-bins, joints $\oplus$ tactile); $W$ belongs to $\mathbb{R}^{d_{out}} * \mathbb{R}^{d_{in}}$, it means that $W$ is a real matrix with dimensions $\mathbb{R}^{d_{out}}$ and $\mathbb{R}^{d_{in}}$.

Notably, the design of each modal data flow path is not parallel fusion, but semantically guided diversion based on task perception requirements, which enables each signal to enter a module with semantic parsing capabilities for structural compression and feature reduction.

### B. Perception Encoder

The perception encoder module receives the raw perception signals output by the sensor ingestion and performs specialized vector encoding based on the multimodal structural characteristics and semantic requirements. The core goal of this process is to establish a set of high-dimensional feature spaces that are normalized and aligned with the task context so that subsequent modules can dynamically expand reasoning in memory iterations. Among the components of this module, the vision backbone receives time series images from visual sensing. It uses a residual convolutional structure and adjustable regularized weights to ensure that high semantic features such as environmental depth, boundary contours, and object relationships are retained while maintaining spatial geometric continuity. The Joint angle MLP receives tensor inputs from proprioceptive and tactile arrays that include joint angular velocity, angular displacement, and end force distribution. It constructs a latent representation of the body's internal dynamic state through nonlinear mapping and feature decomposition using multiple layers of perceptrons. The Audio spectro CNN is a time-frequency convolutional network designed to process acoustic inputs. Its processing flow includes short-time Fourier transform preprocessing and spectrogram convolution operations within a time window. It performs spectral domain decomposition of voice commands, trigger prompts, and environmental reverberations in the sound, extracting rhythm variations and speech semantic cues.

The three sets of modal vectors are combined and feature orthogonalized at the semantic level in fusion dense, and cross-modal dynamic context coupling is achieved by sharing semantic space embedding weights. The algorithm is as follows:

$$f = \tanh(W_f [y_{vis} \| y_{aud} \| y_{pro}])$$

where $\|$ concatenates all modality latents; $W_f$ projects the 224-dim concat into a 256-vector f.

The output of this stage is not just a simple feature fusion, but a primary semantic vector that is used for subsequent reasoning. This vector is not only data passed into the synapse operator, but also marks the formal start of the thinking process. The synapse operator uses the context vector to coordinate the initial state of the neural loop, dynamic timing logic, and tick-slab generation starting point to achieve the transformation process from perception to cognitive reasoning.

C. Tick-Slab CTM Runtime

As a key thinking module in the architecture, the tick-slab CTM runtime is designed to simulate the dynamic memory update, local inference synchronization and multi-path thinking bifurcation of humans in the process of continuous reasoning. The synapse operator at the head end acts as the hub from semantics to neural mapping, performing neural dimension projection on the multimodal semantic vector generated by fusion dense, and converting the tensor into an activation seed that can be interpreted by the neural circuit. The principle is based on the residual concatenation of the semantic tensor and the diachronic memory state, and then generates a composite weight expansion through a shared fully connected layer and a micro MLP module. It ensures that the semantic information has completed semantic normalization and spatial tensor alignment before being translated into tick-slab units. The corresponding algorithm is as follows:

$$\tilde{z}_t = \tanh(W_s [z_{t-1} \| f])$$

where $\tilde{z}_t$ represents the intermediate state vector at the current time step t; $z_{t-1}$ is previous hidden state; f is the fusion vector; $W_s$ map $\mathbb{R}^{D+256} \to \mathbb{R}^D$.

The tick slab fusion unit is the core component of the time recurrence structure. It expands continuous slab units according to the time evolution of thought nodes so that multiple tick-level calculation frameworks are encapsulated in each slab. This unit synchronously expands the semantic projection sequence in a time-increasing manner, and updates the state according to the current context vector, the previous tick state and the coupling of the slab internal parameters. The technical key is that the slab architecture allows horizontal expansion and calculation of trajectories, so that each memory unit can evolve adaptively in an independent tick, thereby breaking through the blocking problem of traditional recurrent models in time series.

The low rank μ-MLP is designed to address the problem of parameter explosion and semantic dilution caused by long-term memory stacking. Its mechanism compresses the semantic space into several μ-level low-rank sub-tensors through kronecker-based tensor decomposition, and uses a shared weight library to selectively and dynamically encode the memory blocks required for each slab. The corresponding algorithm is as follows:

$$z_t^{(d)} = \tanh(b_0^{(d)} + \sum_{m=1}^{M}[\sum_{j=1}^{r} A_{mj} B_{dj}] H_{t,m}^{(d)})$$

where $A \in \mathbb{R}^{M*r}$, $B \in \mathbb{R}^{D*r}$ are shared low-rank factors; $H^d_{t,m}$ is the m-th entry of the depth history buffer for neuron d; $b^d_0$ is a scalar bias.

The low Rank μ-MLP implements compressed back-projection mapping on the computational side to preserve sequence correlation and action intention dimension. At the same time, it suppresses the redundant expansion phenomenon in high-dimensional continuous reasoning, which is the core support for the scalability of the overall memory module.

Sync rank 1 updater is responsible for establishing temporal consistency between slabs, and its principle is derived from neural phase synchronization (Darlow et al., 2025). This module evaluates the rank-1 distribution of the structure vector output by the tick slab, and calculates the dynamic sparse matrix of information convergence in the time domain using the entropy decay model. This design enables each reasoning round to reconstruct the sequential dependencies of past thought evolution and strengthen the coherence of context retention and strategy transfer. The corresponding algorithm is as follows:

$$S_k \leftarrow (\text{decay}^L) S_k + \sum_{j=1}^{L} w_j z_{j,p_k} z_{j,q_k}, \quad w_j = \text{decay}^{L-j}$$

where $S_k$ is k-th synchrony accumulator (one per index-pair ($p_k$, $q_k$)); L ticks in the current slab; $z_{j,*}$ is the hidden state at tick j; decay=0.999.

The updated synchronization matrix then enters the certainty estimator, which generates a differentiable uncertainty vector through the entropy mask and softmax entropy mapping mechanism. The corresponding algorithm is as follows:

$$c = 1 - \frac{H(\text{softmax}(W_c S))}{\log C}, \quad H(p) = -\sum_i p_i \log p_i$$

where S is the P-dim synchrony vector; $W_c \in \mathbb{R}^{C*P}$ (LOGIT_SCALE = 8.0); $C$ = 4 logits.

In the parallel thinking flowchart (Fig. 4), the system uses entropy head as the monitoring module for convergence and overall information entropy change to monitor whether the current slab expansion sequence has reached the thinking termination threshold. After entering the affect epsilon, the system will calculate whether the slab expansion needs to be interrupted based on the variability and synchronization status of the semantic tensor within time. After that, the threshold hold is used as a threshold controller to make a binary decision. If the entropy change and the reasoning signal within the slab are stable, it will be directed to the output writer for semantic output and recording; otherwise, it will enter the slab builder next slab to start the parallel planning of the next slab unit. The core intention of this design is to achieve slab-level time expansion and parallelization.

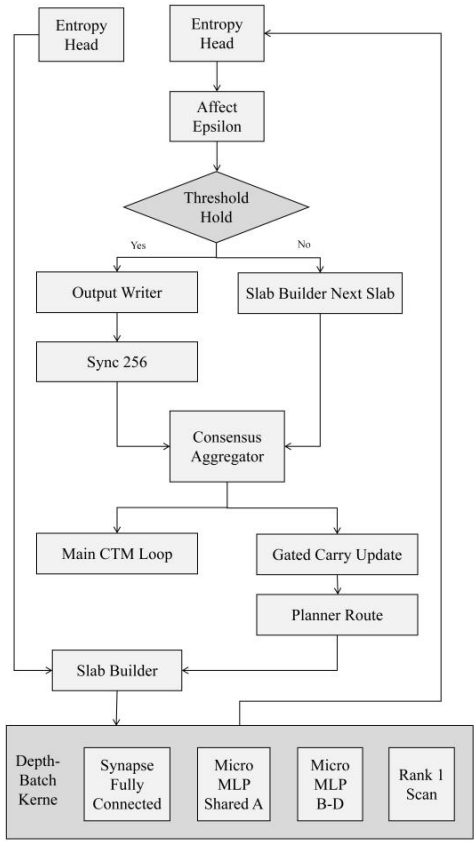

Fig. 4  Parallel workflow of CTM

Slab transfers its internal state to the planner route via gated carry update to execute strategy scheduling and execution logic configuration. It is then sent to the consensus aggregator together with the memory vector generated by sync256 from another branch for vote integration and semantic coordination. The integrated slab's decision output will be distributed to the main CTM loop and gated carry update to maintain the recursive thinking and synchronous recursive operations respectively. Finally, all slabs will be connected to the depth-batch kernel by the slab builder, which contains the synapse FC layer and multi-way micro-MLP structure. The rank 1 scan integrates the semantic focus range of the output nodes to achieve the convergence and reorganization of the semantic parallel path.

D. *Consensus Aggregator*

The consensus aggregator module accepts multiple synchronization vectors and logit structures generated by the certainty estimator. Its purpose is to perform confidence-weighted integration of the results of each CTM slab after the multi-branch parallel reasoning is completed. The Waits for extra slab function is a strategic delay module. Its logic is based on retaining a candidate CTM branch after the main branch returns. And it obtains the synchronization vector with the second highest confidence by forcing an extra slab to wait. Confidence vote merge adopts an entropy-based soft confidence weight mechanism to perform softmax and entropy evaluation on each CTM logit vector. It derives the relative confidence of each candidate vector and uses it as a weighted synthesis coefficient for the synchronization vector, and finally outputs a main synchronization tensor representing the overall consensus of multiple reasoning branches. The corresponding algorithm is as follows:

$$S_{merged} = \frac{\sum_{i=1}^{N} c_i s_i}{\sum_{i=1}^{N} c_i}, \quad c_i = 1 - \frac{H(\text{softmax}(h_i))}{\log C}$$

where $s_i$ is sync vector from branch $i$; $h_i$ is its logits.

The vector will be passed to the affect decoder of the emotional state loop module to drive the emotion modulation loop and dynamically adjust the uncertainty valve, which constitutes the negotiation gate before the perception and reasoning results enter the main control strategy loop.

E. *Emotional State Loop*

The emotional state loop module consists of an affect decoder and a certainty modulator, which is designed to dynamically adjust the trust logic and semantic stability threshold in the process of thought evolution. The affect decoder is responsible for receiving the semantic tensor from the main synchronization vector. It parses the strategic tension and semantic divergence in the context through multi-layer tensor reconstruction and feature attention gating, then transforms it into a situation-oriented dynamic activation signal. The output is used as a semantic marker of the emotional state. On the one hand, it is provided to the serialization module in the MCP envelope router for strategic scheduling; on the other hand, it is handed over to the certainty modulator to adjust the uncertainty control parameters in the cognitive module. The corresponding algorithm is as follows:

$$e_t = \tanh(W_2 \tanh(W_1 S))$$

where $W_1 \in \mathbb{R}^{32*P}$, $W_2 \in \mathbb{R}^{8*32}$; Output is the 8-dim affect vector.

The certainty modulator calculates entropy change and tension difference based on the current context activation and the emotion intensity gradient generated by the previous round of reasoning. It modulates threshold by adjusting the threshold parameter $\varepsilon$ used by the certainty estimator in the tick-slab CTM module. The specific algorithm is as follows:

$$\varepsilon = \varepsilon_0(1 + \alpha \| e_t \|_2), \quad \varepsilon_0 = 0.75, \quad \alpha = 0.5$$

where $\varepsilon_t$ is the uncertainty gain coefficient after emotion modulation at the current time $t$; $\varepsilon_0$ is the initial benchmark confidence scaling factor; $\alpha$ is the modulation sensitivity coefficient; $\|e_t\|_2$ is the L2 norm of the emotion vector $e_t$ at time $t$, it is the total intensity of emotional activation.

The algorithm of gated hidden-state carry (feedback after planner route) is as follows:

$$z_{t+1} = \beta z_t + (1-\beta)\tilde{z}_{t+1}, \ \beta \approx 0.9$$

where $z_{t+1}$ represents the updated state vector; $\beta$ is the smoothing coefficient that controls the proportion of the previous state in the current update; $z_t$ represents the state representation at the current time t, which is the internal vector memory stored in the past time steps; $\tilde{z}_{t+1}$ represents the candidate output calculated at the current moment.

*F. MCP Envelope Router*

The MCP envelope router module is the central node in the architecture that connects the thinking and reasoning results with motion control. It has multi-level control functions such as synchronization encapsulation, strategy diversion, branch competition and fault-tolerant transfer. The serializer formats and encapsulates the synchronization vectors, historical slabs generated by the CTM runtime, and the semantic and emotional information extracted by the affect decoder. It uses structured JSON encoding internally to ensure the semantic consistency and time stability of cross-module transmission. The encapsulation output is the lowest semantic unit of the action command passed to the control layer.

The policy gate makes decisions based on the comparison between the reasoning confidence value estimated by the certainty estimator and the threshold parameter γ. If the confidence value is low, the CTM is directed to execute the next round of thought evolution; if the confidence value is high, the action control process is entered. The parallel branch nanager, based on the branch prediction parallelism emphasized in the architecture, realizes the synchronous thinking competition of multiple CTM clones, and selects the first one to reach the confidence threshold as the main decision response. If no branch is reached, the result of the last completion is taken.

The timeout safe pass module is the core of the fault-tolerant design. When any branch fails to meet the stop condition within a certain time, the module will use the previously cached synchronization vector to automatically trigger the action output process and enter the actuator controller. From a technical perspective, the output vector directly enters the trajectory planner in the actuator controller module to fill the gap when the planner module is blocked for a long time or fails unexpectedly. The principle ensures that the system maintains stable operation and prevent non-output interruption action in low-confidence and high-latency scenarios.

*G. Actuator Controller*

The actuator controller is the translation hub from semantic decision to physical execution, which means it is responsible for the layer-by-layer mapping from semantic drive to control signal. The trajectory planner is responsible for deconstructing the synchronization vector into the correspondence between high-level behavioral semantics and spatial targets, and internally adopts timestamp-oriented joint path interpolation logic. According to the action phase and control node represented by the vector tensor, a continuously differentiable reference displacement curve is generated. It ensures that the action output is no longer a discrete point, but a time-consistent action trajectory. The algorithm is as follows:

$$\min_{\tau} \| \tau - K\, s_{merged} \|_2^2, \ \text{s.t.}\ \tau_{min} \leq \tau_i \leq \tau_{max}$$

where $\tau$ is the output torque vector to be optimized; $K$ is a constant mapping matrix; $s_{merged}$ represents the expected input of the merged state; $\tau_{min}$ / $\tau_{max}$ are the upper and lower limits of the torque for each joint.

The compliance filter mainly uses analog compliance compensation, and then processes the above trajectory output and suppresses the unstable high-frequency components in the path signal to prevent the rigid drive from causing reverse interference to the structure. The filter combines forward differential and sliding average operations internally, so that the control output has both real-time and smoothness, which is especially suitable for continuous joint control scenarios. The torque mapper finally maps the motion trajectory after compliance processing to the corresponding control voltage or PWM output range. Its core logic is the segmented linear mapping table and gain factor adjustment mechanism to ensure that each channel signal can be converted into a low-level execution signal with physical consistency.

IV. EXPERIMENTS

Given research onion, this work adopts the philosophical paradigm of positivism that stems from the nature of the research object (Saunders et al., 2009). In terms of experimental methods, the researcher chose to use a simulation-based experiment as the main empirical means. This method uses a high-fidelity virtual environment to simulate real operating conditions to verify the validity and stability of the internal thinking process and action strategy of the humanoid control system in a specific situation (HSaglam & Papelis, 2024). It is particularly suitable for systems involving high-degree-of-freedom control outputs, multi-stage semantics and signal translation processes that require observation, and systems that cannot be directly observed by human operation (Li et al., 2024). Specifically, since the actuator controller module of this architecture already covers from semantic vectors to actual control signals. It may be difficult to verify physical consistency and signal stability using only instruction-level testing in real experiments, which makes it difficult to analyze the architecture's ability to adjust to real mission scenarios.

Simulation-based experiments facilitate control comparison of strategy selection and semantic evolution paths without being interfered by hardware limitations, thereby providing behavioral-level explanation capabilities far superior to static verification (Saglam & Papelis, 2024). In addition, without simplifying the architecture logic and data flow model, the simulation environment can fully support the closed structure of CTM-MCP. It allows researchers to observe the state changes of the entire path from semantic thinking to action output, thereby constructing a verified physical action mapping logic (Li et al., 2024).

## A. Experimental Setup

The necessity of choosing o4-mini-high as the core execution platform for simulation experiments is not due to its universality as a replacement for traditional control tools, but rather due to the architecture's need for structural synchronization of cross-layer data flows between semantics, reasoning, decision-making, and control. Current tools for building robot control systems, such as ROS, Gazebo, or Isaac Sim, focus more on kinematics solving, sensor modeling, and underlying physical simulation (Salimpour et al., 2025). However, this work requires support for continuous thinking and evolution at the semantic level, such as the tick-slab recurrent memory constructed by CTM and the need for encapsulated semantic synchronization operations between multiple modules. In addition, the package vector generated by MCP needs to go through continuous semantic judgment and confidence vote before entering the control signal mapping layer, which has to rely on the language layer reasoning model to complete slab expansion, vector sorting, threshold update and other logic.

The researcher designed and wrote the architecture code of each module and the experimental code using Python 3.13 IDLE according to the principle of CTM and MCP in this work, which is used to run on o4-mini-high. The designed experimental group code follows the tick-slab CTM runtime, MCP envelope router and other modules to build the experimental environment throughout the process. Since the latest version of ChatGPT supports the function of dynamically running code, the researcher can run the entire process of module logic, verify semantic flow delivery, encapsulation decision and control signal mapping, and then simulate the complete "thinking-planning-execution" of the humanoid robot. The control group selected NVIDIA's open-source Isaac GR00T code, whose thinking, reasoning, planning, and cognitive control code structure is open on GitHub. To ensure that the control group code and the experimental group eliminate the hidden dangers of unfairness during operation, the researcher adjusted the original code format and calling order while retaining the original functional design of NVIDIA Isaac GR00T. The architecture code and experimental code are uploaded to the Github repository and are publicly available.

## B. Dataset

This work uses SayCan proposed by Ahn et al. (2022) of the Google DeepMind team as the semantic task dataset. The dataset is designed to integrate language model reasoning and the action planning framework of robot operation decision-making. 100 task examples released by the official are widely used in LLM-to-robot action generation research (Ahn et al., 2022). It accurately describes the multidimensional properties of the robot's language-to-action translation process, and corresponds to the CTM-MCP architecture in semantic continuous reasoning and encapsulated decision-making structure. Compared with other task datasets such as ALFRED, BEHAVIOR or TEACh, the semantic task structure provided by SayCan just meets the needs of tick-slab thinking and semantic synchronization vector update within the slab.

Considering that the original SayCan only has 100 open-source tasks, the researcher synthesized an additional 900 tasks based on the semantic arrangement of the tasks, the scope of operational variables, and the logical flow of the tasks. These synthetic data were verified through performance testing and semantic stability to ensure that they are consistent with the original samples in terms of function and logical action. The process of synthesizing data follows the contextual internal structure of the original data, retaining the logical hierarchy of "goal-operation path-expected response" in each task to ensure semantic coherence and reasoning universality.

## C. Implementation

The researcher first passed the framework code and experimental code designed and tested by Python 3.13 IDLE into the OpenAI o4-mini-high model to build an experimental environment as the operating core of the semantic framework simulation environment. The framework code is responsible for initializing various modules, encapsulation nodes and control translation layers such as tick-slab CTM runtime and MCP envelope route. The experimental code builds the semantic perception and action simulation environment. Through continuous context maintenance and state progression during the operation of o4-mini-high, a semantic simulation covering the "thinking-planning-action" closed-loop logic is gradually formed. Subsequently, the researcher imported the expanded and synthesized SayCan dataset into the control system simulated by o4-mini-high to start the verification of the structured semantic task execution. To construct a control group benchmark, the researcher input the formatted and sequence-aligned NVIDIA Isaac GR00T code in another independent o4-mini-high dialogue session. The same SayCan dataset was then imported to ensure that the semantic processing and logical delivery paths between modules were structurally consistent and comparable.

Considering the uncertainties such as execution interruption, unstable transmission and memory conflict that may occur during the simulation experiment, this work repeats the entire experimental process for 9 epochs. In each epoch, multiple debugging, restart and repair operations are performed for abnormal execution and erroneous logic. Finally, only a set of results with the most stable performance and the most complete task logic are selected for recording. The experimental records have been uploaded to the GitHub repository for open access to ensure data credibility and experimental traceability.

In addition, during the data analysis phase, the researcher used seven metrics: task success rate (TSR), execution success rate (ESR), average episode length (AEL), ROSCOE, REVEAL, proficiency self-assessment (PSA), and task effectiveness (TE).

## V. RESULT & DISCUSSION

As the experiment was completed and recorded, this work used the mathematical formulas based on the seven evaluation indicators to calculate the data results of the experimental group and the control group in the o4-mini-high model. OpenAI's o4-mini-high model uses high-frequency memory maintenance and powerful computing power to stably convert the execution results generated by each epoch experiment into quantifiable values, which improves the objective reliability and validity of this study. Table 1 and table 2 respectively present the execution results of the CTM-MCP simulation control system and the NVIDIA Isaac GR00T control system in the nine epochs simulation experiment as a basis for performance comparison.

Table 1     The metrics for the CTM-MCP

| Epochs | TSR | ESR | AEL | ROSCOE | REVEAL | PSA | TE |
|---|---|---|---|---|---|---|---|
| 1 | 0.902 | 0.799 | 10.152 | 0.787 | 0.829 | 0.882 | 0.859 |
| 2 | 0.876 | 0.781 | 10.223 | 0.780 | 0.777 | 0.878 | 0.857 |
| 3 | 0.894 | 0.803 | 9.412 | 0.789 | 0.782 | 0.849 | 0.855 |
| 4 | 0.901 | 0.795 | 10.486 | 0.788 | 0.826 | 0.874 | 0.854 |
| 5 | 0.891 | 0.777 | 10.032 | 0.779 | 0.773 | 0.864 | 0.839 |
| 6 | 0.906 | 0.801 | 9.957 | 0.785 | 0.823 | 0.884 | 0.858 |
| 7 | 0.893 | 0.801 | 10.128 | 0.784 | 0.783 | 0.889 | 0.861 |
| 8 | 0.888 | 0.794 | 9.681 | 0.786 | 0.822 | 0.873 | 0.856 |
| 9 | 0.911 | 0.803 | 9.875 | 0.791 | 0.783 | 0.887 | 0.861 |

Table 2     The metrics for the NVIDIA Isaac GR00T

| Epochs | TSR | ESR | AEL | ROSCOE | REVEAL | PSA | TE |
|---|---|---|---|---|---|---|---|
| 1 | 0.660 | 0.732 | 9.536 | 0.655 | 0.675 | 0.715 | 0.786 |
| 2 | 0.652 | 0.724 | 9.620 | 0.667 | 0.689 | 0.721 | 0.787 |
| 3 | 0.667 | 0.725 | 9.477 | 0.661 | 0.683 | 0.721 | 0.780 |
| 4 | 0.657 | 0.737 | 9.637 | 0.666 | 0.680 | 0.722 | 0.789 |
| 5 | 0.668 | 0.727 | 9.593 | 0.664 | 0.681 | 0.720 | 0.782 |
| 6 | 0.658 | 0.732 | 9.510 | 0.667 | 0.684 | 0.724 | 0.776 |
| 7 | 0.655 | 0.740 | 9.592 | 0.667 | 0.687 | 0.728 | 0.787 |
| 8 | 0.652 | 0.734 | 9.470 | 0.654 | 0.688 | 0.728 | 0.776 |
| 9 | 0.652 | 0.737 | 9.474 | 0.661 | 0.675 | 0.721 | 0.777 |

The researcher then used a line graph to show the changing trends of seven evaluation indicators of the control system simulated by CTM-MCP and NVIDIA Isaac GR00T in nine rounds of simulation experiments (Fig.5). This visual approach helps to observe the performance fluctuations of each metric and the quantitative evidence of the operating bottleneck, which supports the conclusions drawn in stability comparison and system resilience analysis.

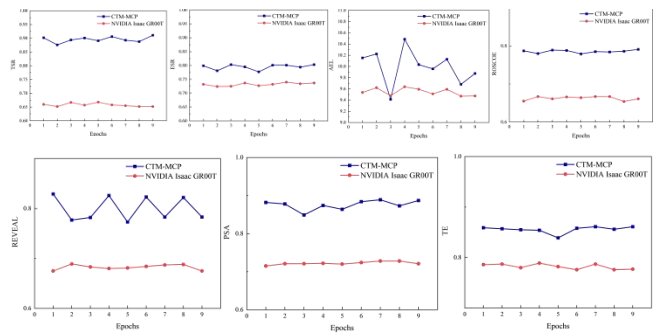

Fig. 5     Comparison of the trends of 7 metrics between CTM-MCP and NVIDIA Isaac GR00T

In the results of 9 epochs, CTM-MCP showed higher maximum values and lower standard deviations in all seven metrics, which means excellent performance and high stability. In terms of TSR, its highest value is 0.911, and the standard deviation is only 0.0105. It is much better than NVIDIA Isaac GR00T's 0.668 and 0.0062. In terms of ESR, CTM-MCP reached 0.803, while Isaac GR00T only reached a maximum of 0.740. In the AEL part, CTM-MCP reached a maximum of 10.486, which means that it is more detailed in task analysis and action selection. Although the volatility is high, it is explained by the complexity of the task and the flexibility of the path; the maximum value of the control group is only 9.637 and the volatility is very small, which shows that the behavior execution is relatively rigid. Analysis of the two composite cognitive representation metrics ROSCOE and REVEAL revealed that CTM-MCP achieved the highest values of 0.791 and 0.829 respectively and remained stable; Isaac GR00T was 0.667 and 0.689, and the fluctuation was significantly larger. PSA and TE also showed consistent trends. The maximum value of CTM-MCP was 0.889, which indicated that the module had strong self-assessment ability; the latter was 0.861, which reflected the high consistency of overall task performance. The relative value of Isaac GR00T was significantly lower, with the highest values of PSA and TE being 0.728 and 0.789 respectively.

The line graph verifies the stable performance of CTM-MCP on various indicators. Except for the predictable high and low fluctuations of AEL, other indicators such as TSR and ROSCOE all show a stable upward or flat trend, showing that this architecture has high reliability and strong resilience. In contrast, the Isaac GR00T curve changes more irregularly, showing problems such as large volatility and insufficient accuracy. The above line graph results consolidate the systematic advantages and stable structure of CTM-MCP in multi-indicator evaluation. Through the cross-analysis of objective data and images, the credibility and universal potential of CTM-MCP are strengthened.

## VI. LIMITATION & FUTURE RESEARCH

Given that the simulations were run with OpenAI's o4-mini-high model, verifying CTM's parallel acceleration in practice would require replacing list loops with fused CUDA or Triton kernels. The current architecture code and experimental code simulate logic but cannot achieve responsive acceleration based on compute power. This is acceptable for research demonstrations, but for product prototypes, the learned parameters need to be loaded. The latency and quota limitations of LLMs may mask CTM's millisecond adaptive computing advantage, which limits the real-time reflection of the measurement environment to the humanoid robot control system. If end-to-end timing is performed through the API, CTM's millisecond pause advantage is less significant.

Future researchers will need to run locally to achieve time parity. Local CPU/GPU fallback scripts will also be needed for timing, so that the results do not depend on OpenAI's queue delays. In addition, this work only studies "thinking-planning-execution" based on clear semantics, and does not involve data multimodality in the perception field. In other words, the state of the art in the field of sensors and multimodality of humanoid robots will affect whether clear semantics can be obtained.

## VII. CONCLUSION

In response to the long-standing cognitive and behavioral disconnect in humanoid robots, this work proposes a CTM-MCP architecture for use in control systems. It addresss the problem that traditional humanoid robots have difficulty in "thinking-planning-acting" in real time for a single specific purpose in unfamiliar scenarios. It provides a solution to the current situation based on static preset rules and engineer-instruction coding, which lacks the ability to reorganize sensory inputs in real time and dynamically reason. In addition, CTM-MCP addresss the current situation that humanoid robots are difficult to autonomously encode in time to achieve autonomous actions due to the highly programmed "call tool-return result" model. This architecture contributes to solving the above-mentioned technical gap. Through continuous thinking, MCP operating tools are driven by perception dynamic modeling and semantic translation control to achieve truly autonomous coding and self-driven actions.

The tick-slab CTM runtime converts the original sequential reasoning operation into a parallelizable module process through Tick-Slab fusion and rank compression mechanism, breaking through the limitations and proposing a specific parallel solution in CTM theory. It cooperates with the synchronous rank-one updater and low-rank μ-MLP to suppress the problem of lengthy deep parameters, so that the model no longer relies on large-scale parameter stacking. The MCP envelope router structures the high- and low-confidence policy gating and branch management mechanisms to achieve the encapsulation and operation translation of semantic decisions, and ultimately the corresponding trigger action control is executed by the execution module.

In the simulation-based experiments, this work uses o4-mini-high as the execution platform and completes nine epochs of cross-comparison tests. Compared with the control group NVIDIA Isaac GR00T, the experimental group CTM-MCP shows continuous leadership and stability in seven metrics on the SayCan dataset task accuracy and execution stability indicators. Overall, the CTM-MCP architecture breaks through the logical bottleneck of existing models, enabling humanoid robots to have the closed-loop ability to drive internal semantic thinking with scene perception, and then extend to timely action planning and dynamic operation. It provides a reference experience for autonomous bodies to get rid of external artificial coding and move towards internal self-coding of the scene to achieve human-like dynamics and autonomy.